\documentclass{ieeeaccess}
\usepackage{cite}
\usepackage{amsmath,amssymb,amsfonts,bm}
\usepackage{algorithm}
\usepackage{algorithmic}
\usepackage{hyperref}
\usepackage{graphicx}
\usepackage{textcomp}
\def\BibTeX{{\rm B\kern-.05em{\sc i\kern-.025em b}\kern-.08em
    T\kern-.1667em\lower.7ex\hbox{E}\kern-.125emX}}
\begin{document}
\history{Date of publication xxxx 00, 0000, date of current version xxxx 00, 0000.}
\doi{10.1109/ACCESS.2017.DOI}

\title{Bayesian graph convolutional neural networks via tempered MCMC}
\author{\uppercase{Rohitash Chandra *}\authorrefmark{1} \IEEEmembership{SM, IEEE},
\uppercase{Ayush Bhagat * \authorrefmark{2}, Manavendra Maharana \authorrefmark{2}, Pavel N. Krivitsky \authorrefmark{1}}
 }
\address[1]{School of Mathematics and Statistics, UNSW Sydney, Kensington, NSW 2052, Australia (e-mail: rohitash.chandra@unsw.edu.au)}
\address[2]{Manipal Institute of Technology, Karnataka, India}
\tfootnote{* Authors contributed equally.}

\markboth
{Chandra et al. \headeretal: IEEE Access}
{Chandra et al.  \headeretal: IEEE Access}

\begin{abstract}

Deep learning models, such as convolutional neural networks, have long been applied to image and multi-media tasks, particularly those with structured data. More recently, there has been more attention to unstructured data that can be represented via graphs. These types of data are often found in health and medicine, social networks, and research data repositories. Graph convolutional neural networks have recently gained attention in the field of deep learning that takes advantage of graph-based data representation with automatic feature extraction via convolutions. Given the popularity of these methods in a wide range of applications,  robust uncertainty quantification is vital. This remains a challenge for large models and unstructured datasets. Bayesian inference provides a principled approach to uncertainty quantification of model parameters for deep learning models. Although Bayesian inference has been used extensively elsewhere, its application to deep learning remains limited due to the computational requirements of the Markov Chain Monte Carlo (MCMC) methods. Recent advances in parallel computing and advanced proposal schemes in MCMC sampling methods has opened the path for Bayesian deep learning. In this paper, we present Bayesian graph convolutional neural networks  that employ  tempered  MCMC sampling with Langevin-gradient proposal distribution implemented via parallel computing.  Our results show that the proposed method can provide accuracy similar to advanced optimisers while providing uncertainty quantification for key benchmark problems. 
 
\end{abstract}

\begin{keywords}
Bayesian neural networks; MCMC; Langevin Dynamics; Bayesian deep learning; and Graph neural networks.
\end{keywords}

\titlepgskip=-15pt

\maketitle

\section{Introduction}
\label{sec:introduction}
Graph neural networks are a type of artificial neural network designed for data which features graph-based representation \cite{scarselli2008graph,wu2020comprehensive,zhang2020deep,xu2018powerful}. Graph-based representation can be used to analyse non-structured and non-sequential data, such as a social network comprising users and their activities \cite{cook2006mining}. Recently, a wide variety of graph-based deep learning network architectures has been introduced \cite{zhang2020deep}, such as graph convolutional neural networks (CNNs) \cite{gao2018large, wu2019simplifying, schlichtkrull2018modeling}, graph recurrent neural networks featuring long short-term memory (LSTM) networks \cite{liang2016semantic,zayats2018conversation,tang2019coherence,shu2020host}, graph auto-encoders \cite{wang2017mgae,pan2018adversarially}, and graph generative adversarial networks (GANs) \cite{wang2017graphgan}.  
Applications of graph neural networks have included time series forecasting \cite{wu2020connecting}, traffic flow forecasting \cite{chen2019gated,peng2020spatial}, particle physics \cite{shlomi2020graph}, molecular property prediction  \cite{wieder2020compact}, sentiment analysis \cite{wang2020novel}, recommender systems \cite{wang2019knowledge}, and social media popularity prediction \cite{cao2020popularity}. A review  of  applications of graph neural networks  has been given in \cite{zhou2018graph}.

Deep learning methods, such as convolution neural networks  \cite{lecun1998cnn,russ2015alexnet} (CNNs) and recurrent neural networks  \cite{hochreiter1997long,schmidhuber2015deep} (RNNs) have been applied to image data and temporal sequences; however, these are structured, regular, Euclidean data, although  they can be viewed as graphs (i.e., lattices). CNNs and RNNs are less applicable to unstructured or graph-based data with multi-layer hierarchical structure, with features that occur on different scales. On the other hand, graph neural networks (GNNs) use graph-based representation of data to propagate on each node. Aspects of the data such as the input order of the nodes are irrelevant, with the graph instead representing the  dependencies between them; hence, GNNs can enable propagation guided by graph structure as done in   simple (canonical)  neural networks \cite{Yang2016,Kipf2016}.  

As the impact of graph neural networks on different deep learning architectures and applications grows, there is also a growing need for robust uncertainty quantification in model parameters. Bayesian inference provides a means for robust uncertainty quantification in deep learning models by sampling from the posterior distribution that represents the model parameters \cite{neal1996sampling,welling2011bayesian} using  Markov-Chain Monte Carlo (MCMC) methods \cite{CHANDRA2020EAI,chandra2019langevin_}.  Implementation of Bayesian inference via MCMC becomes very challenging  with growing  size of the data and the number of model parameters. MCMC methods have extensive computational requirements as thousands or samples needed for training, and hence limited work has been done in  implementing deep learning models such as  Bayesian CNNs via MCMC sampling. Variational inference, provides an alternative for uncertainty quantification in deep learning methods via \textit{Bayes by backpropagation} \cite{blundell2015weight}. At the same time, there has been progress in the MCMC approach, making use of  parallel computing and advanced MCMC methods \cite{CHANDRA2020EAI,chandra2019langevin_}, which can enable the framework for  graph CNNs. Advanced proposal distribution in MCMC that incorporate gradients \cite{chandra2019langevin_, welling2011bayesian,chandrakapoor2020bayesian} has opened the door to Bayesian deep learning methods for novel deep learning methods.

In this paper, we present Bayesian graph convolutional neural networks that employs  tempered  MCMC sampling via parallel computing with   Langevin-gradient proposal distribution. We  apply the method to selected benchmark graph-based datasets obtained from research data repositories such as \textit{PubMed} \footnote{\url{https://pubmed.ncbi.nlm.nih.gov/}}. The  parallel computing framework features inter-process communication for exchange of tempered MCMC replica states as  demonstrated previously for Bayesian neural networks \cite{chandra2019langevin_}.

The remainder of the paper is organised as follows. In Section~\ref{sec:background}, we review the background of the problem and related methods. Section~\ref{sec:methodology} presents the proposed methodology, followed by experiments and results in Section~\ref{sec:results}. We discuss the implications of our work and directions of future work in Section~\ref{sec:discussion},  and Section~\ref{sec:conclusion} concludes the paper.

\section{Background and Related Work}
\label{sec:background}

\subsection{Graph neural networks}

A graph $G$ data structure consists of a  set of vertices (nodes) $V$ and edges $E$, which can be either directed or undirected \cite{bondy1976graph,west2001introduction}. Each node represents a data element, and the edges denote the relationships between the data elements. Each node has its own graph embedding via a feature vector, which summarises the properties of that particular data element. The nodes send their graph embedding to their immediate neighbours in the form of messages \cite{Gilmer}. 

The message received by node $v$ at GNN layer index $t$, $m_{v}^{t}$ is constructed by aggregating over the set of neighbours $v$, $N(v)$, the the results of a message function $M_{t}$ which takes three arguments: the feature vector of $v$ itself ($h_{v}^{t}$), the feature vector of neighbour $w$ ($h_{w}^{t}$), and the features of the edge between $v$ and $w$ ($e_{vw}$):
\begin{equation}
m_{v}^{t}=\sum_{w\in N(v)}M_{t} (h_{v}^{t},h_{w}^{t},e_{vw}).
\label{eqn:msggraph}\end{equation}
Based on this message $m_{v}^{t}$ and the previous value $h_{v}^{t}$, the latter is updated via a function $U_{t}$:
\begin{equation}
h_{v}^{t+1}=U_{t}\,(h_{v}^{t},m_{v}^{t}).
\label{eqn:updategraph}
\end{equation}

A variety of graph representations are possible, including directed graphs, heterogeneous graphs,  edge information graphs, and dynamic graphs  \cite{zhou2018graph}. The goal  of GNN is to learn a state
embedding  which incorporates the neighbourhood  information of
 for each node; this state embedding can then be used for classification and other purposes. The choice of the propagation and the output step of a GNN are required to obtain the hidden states of nodes or edges and depend on the application. In relation to the canonical GNNs, the focus has been on refinements in the propagation step, while a simple feedforward neural network is retained in the output step. The major variants in the propagation step utilise different aggregators to gather information from each node’s neighbors. Some of the key propagation step methods include attention aggregator (graph attention network \cite{gan2018} and gated attention network \cite{DBLP:journals/corr/abs-1803-07294}), gated aggregator (gated graph neural networks \cite{li2017gated} and graph LSTM \cite{graphlstm2018}), skip connection (highway GNN \cite{2018highway} and jump-knowledge network \cite{jump2018}), hierarchical graph   edge conditioned convolution  \cite{ECC2017}), and finally convolutional aggregator, which features graph CNN with spatial \cite{NIPS2016_390e9825} and spectral methods \cite{Kipf2016}. 
 
 Spatial methods in graph CNNs include neural fingerprints (FPs) \cite{DuvenaudMAGHAA15}, dual graph convolution network (DGCN) \cite{dgcn}, and model networks (MoNet) \cite{MoNet}. Some of the commonly used spectral methods in graph CNNs are ChebNet (Chebyshev polynomial approximation algorithm) \cite{HAMMOND2011129}, graph convolutional networks (GCNs)  which is first-order approximation of graph convolutions  \cite{Kipf2016},  and adaptive graph convolutional networks (AGCN) \cite{AGCN}. In spectral convolution, the underlying structure of the graph is deduced by eigen-decomposition of the graph laplacian. The entire graph is processed simultaneously, which makes spectral convolution more computationally expensive. However, it is still widely used, because the spectral filters excel at capturing complex patterns. Spatial methods use information from neighbouring nodes and deduce properties of a node based on features of its closest  $k$ neighbours. The graph can be processed in batches of nodes, which help improve speed and efficiency.
 
 A number of approaches can be used to update or training GNNs. They include neighbourhood sampling (Graph SAGE \cite{graphsage2017}, Pin-Sage \cite{pinsage2018} and Fast-GCN \cite{fastgcn2018}), receptive field control \cite{chen2018stochastic}, data augumentation such as co-training   and self-training \cite{gao2018large}, and unspervised training such as graph autoencoder (GAE) \cite{kipf2016variational} and adversarially regularized graph autoencoder (ARGA) \cite{pan2019adversarially}.
 
 \subsection{Bayesian Deep Learning}
 %xx
 
 Research in area of Bayesian deep learning has been limited due to the limitations of canonical MCMC methods for large number of parameters, and other characteristics  such as complex   architectural properties of  deep learning models \cite{wang2020survey,polson2017deep}. As we noted earlier, variational inference provides a computationally cheaper approach, with variational autoencoders \cite{kusner2017grammar}, variational autoencoders and GANs \cite{mescheder2017adversarial}, pruned variational CNNs \cite{zhao2019variational,chien2017variational,zhou2020variational},   variational RNNs and long short-term memory (LSTM) networks \cite{su2018variational}, recurrent variational graph convolutions \cite{bonner2019temporal},  and variational graph neural networks with focus on autoencoders \cite{kipf2016variational} and Markov networks \cite{qu2019gmnn}. Most of these methods were developed  in the last five years, particularly after 2017, with applications summarised in \cite{wang2020survey}. Although work has been done in area of variational CNNs \cite{gal2015bayesian}, we did not find any work in the area of variational graph-CNNs. 
  
Over the last decade, advanced proposal distributions incorporating gradients have been applied, such as  Langevin and  Hamiltonian MCMC for statistical models \cite{welling2011bayesian,neal2011mcmc}. However, only in last five years has there been progress in area of Bayesian neural networks with Hamiltonian MCMC   \cite{levy2017generalizing}, and  graphic processing unit (GPU) implementation to enhance computation \cite{cobb2020scaling}.  Langevin MCMC methods   for neural networks include the use of tempered (parallel tempering)  MCMC for simple neural networks applied to pattern classification and time series prediction problems \cite{chandra2019langevin_}. Furthermore,  surrogate assisted estimation via Langevin tempered MCMC has been developed for Bayesian neural networks which is useful when the model and data are computationally expensive   \cite{chandra2020surrogate}. Transfer learning has   been used to take advantage of multiple sources of data in a Bayesian framework via Langevin MCMC sampling  \cite{chandra2020bayesian}. Although simple neural networks have been used in pattern classification problems \cite{chandra2020surrogate,chandra2020bayesian}, some of the problems  had large numbers of features, and hence the neural network models had more than 5,000 parameters, which is comparable to smaller deep learning models.

\section{Methodology}
\label{sec:methodology}

In this section, we present details for Bayesian graph convolutional neural networks  (Bayes-GCNN), which uses spectral convolution  and  tempered  MCMC sampling framework   with Langevin-gradient proposal distribution. The framework is used  for classification of nodes in datasets with graph representation. 

\subsection{Model and likelihood function}

In conventional CNNs, the input data are multiplied by a matrix of weights having the same dimension as the input data. This matrix of weights is known as the \emph{filter} or \emph{kernel}. A single layer can have multiple filters to extract different features in the data. The output can then be fed into more convolution layers or pooling layers to extract the prominent features of the data. Given a CNN  with multiple layers, the output $Y_i^{l}$ for layer $l$ with $m_1$ filters is computed as follows
\begin{equation}
Y_{i}^{(l)}=B_{i}^{(l)}+\sum_{j=1}^{m_{1}^{(l-1)}}K_{i,j}^{(l)}X_{i}^{(l-1)} 
\label{Convolutional Formula}
\end{equation}
where, $B_i^{(l)}$ is the bias matrix, $X_i^{(l-1)}$ is the input data to the layer, and $K_{i,j}^{(l)}$ is the filter.

In comparison to conventional neural networks, graph neural networks do not operate on euclidean data, with a fixed dimension and a tabular structure. Graph information such as the edge directionality, node attributes, and edge attributes cannot necessarily be mapped to a higher dimension euclidean space. In our proposed Bayesian graph CNN (Bayes-GCNN), we use the fast approximate spectral graph convolution technique of Kipf et al. \cite{Kipf2016}. We adapt Equation \eqref{Convolutional Formula} such that it can work on graphs, where the number of node connections is dynamic and the nodes are unordered. Hence,  the equation to compute the convolved signal matrix is given as follows 
\begin{equation}
Z=D^{(-1/2)}AD^{(-1/2)}X\theta 
\label{Graph Convolutional Formula}
\end{equation}
where $Z$ is the convolved signal matrix, $D$ is the degree matrix, $A$ is the adjacency matrix, $X$ is the node feature vector and $\theta$ is a matrix of filter parameters.

\begin{figure*}[htpb!]
 \centering
  \includegraphics[width=15cm]{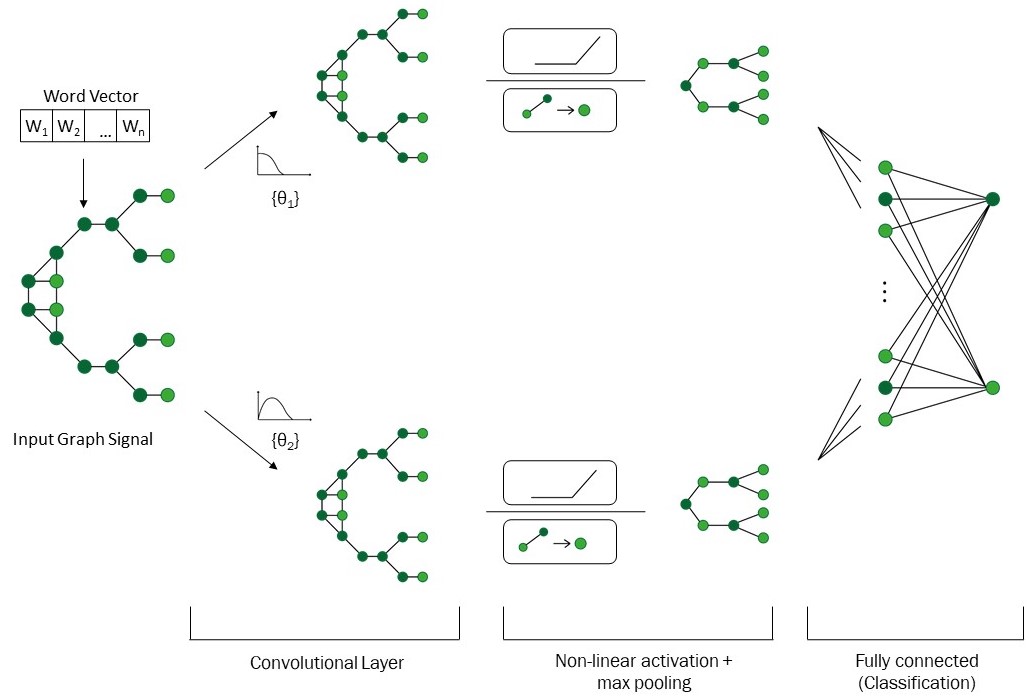}
 \caption{Graph convolutional neural network (GCNN) showing convolutional and pooling layers. }
\label{fig:gcnn}
 \end{figure*}

Our focus in this paper is the application of Bayes-CNN to node classification using graph datasets. Therefore,  we construct the likelihood function that is  used for MCMC sampling of parameters (weights and biases) in Bayes-GCNN. The likelihood function enables the comparison of the training  data featuring graph representation $\mathbf{g}$ with graph CNN output $\mathbf{g'} = z(\mathbf{g}, w, b)$; where, $w$ and $b$ are the Bayes-GCNN parameters (weights and biases combined)  shown in Figure~\ref{fig:gcnn}, respectively. These weights and biases include those in convolution and max-pooling layers of graph CNNs.

Pattern classification problems entail discrete outcomes, and thus we use  a multinomial likelihood function. Suppose that we have  $K$ classes in the training data, and assume that the outputs $\bm{y}=(y_1,\dotsc,y_N)$ are drawn from a multinomial distribution with parameters $\bm{\theta}=(\theta_1,\dotsc,\theta_K)$, for $\sum_{k=1}^K\theta_k=1$. We define indicator variables
\begin{equation}
z_{i,k} =
\begin{cases}
    1,& \text{if $y_i = k$}\\
    0,& \text{otherwise}
\end{cases}
\end{equation}
for observations $i=1,\dotsc, N$ and classes $k=1,\dotsc,K$. Then, the multinomial likelihood function can be expressed as
\begin{equation}
p(\bm{y}|\bm{\theta}) =\prod_{i=1}^N \prod_{k=1}^K \theta_{i,k}^{z_{i,k}},
\label{multinomial}
\end{equation}
where $\theta_{i,k}$ is the Bayes-GCNN model's predicted probability that observation $i$ is in class $k$. We use a multinomial expit (softmax) function to link the output $f(x_i)$ (of inputs $x_i$) of the Bayes-GCNN to the predicted probability:
\begin{equation}
    \theta_{i,k} = \frac{\exp(f_k(x_i))}{\sum_{j=1}^K \exp(f_j(x_i))}.
\end{equation}
The  prior distribution  is given by
\begin{equation}
p(\bm{\theta}) 
    = \frac{1}{(2\pi\sigma^2)^{P/2}} \times \exp\Bigg\{-\frac{1}{2\sigma^2}\bigg(     
  \sum_{i=1}^P \theta_i^2\bigg) \Bigg\}
\label{prior_class}
\end{equation}
where  $\bm{\theta}$ represents the GCNN parameters (weights and biases), $P$ is the total number of parameters, and $\sigma^2$ is user-defined variance which is typically obtained from   prior knowledge about distribution of parameters  in trained neural networks.

Our implementation performs these calculations on the log scale to minimise numerical instability.

 \subsection{Langevin-gradient proposal distribution }

Next,  we use MCMC sampling to sample the posterior distribution of weights and biases of Bayes-GCNN. Deep learning models (such as GCNNs),   can feature   hundreds of thousands of parameters,  hence state-of-art MCMC sampling methods are needed. Therefore, we employ 1) efficient proposal distribution via  Langevin-gradients, 2) parallel computing that features inter-process communication, and  3)  tempered MCMC to optimise sampling from multi-modal posterior distributions.  

The Langevin-gradient  proposal distribution essentially incorporates Gaussian noise with a gradient step taken using   a single iteration (epoch) \cite{welling2011bayesian}. The gradient step  can be either in form of stochastic-gradient descent (SGD) or adaptive gradient-descent, such as the  Adam optimiser \cite{kingma2014adam,chandra2019langevin_}. Henceforth, we refer to  SGD-based proposal distribution as Langevin-gradients (LG) and Adam-based proposal distribution as adaptive Langevin gradients (adapt-LG). 

At a given step or   chain position ($n$)  of a  MCMC sampler, we create a proposal denoted by superscript ($^\star$) from a multivariate normal distribution $\bm{\theta}_{n}^{\star}$ as follows:
\begin{equation}
 \bm{\theta}_{n}^{\star}\sim\mathcal{N}(\bm{\theta}_n+\nu_1\bar{v}_{n}, \nu_2^2 I_P).\label{update}
\end{equation}
Here, $I_P$ is an $P \times P$ identity matrix, scaled by the tuning parameter $\nu_2^2$; and $\bar{v}_{n}$ is the Langevin-gradient $\nabla_\theta \log \{p(\bm{\theta})p(\mathbf{x}|\bm{\theta})\}$ \eqref{prior_class} and \eqref{multinomial}, respectively, scaled by $\nu_1$, either fixed in the case of SGD-based algorithm or adaptive in the case of an Adam-based algorithm \cite{kingma2014adam}. Thus, the proposal attempts to ``climb'' up the posterior density. The Metropolis--Hastings probability ($\alpha$) is  used to accept/reject a proposed sample is as follows
\begin{equation}
\alpha = \min\bigg\{1, \frac{p(\textbf{x}|\bm{\theta}_{n}^{\star})p(\bm{\theta}_{n}^{\star})Q(\bm{\theta}_{n}|\bm{\theta}_{n}^{\star})}{p(\textbf{x}|\bm{\theta}_{n})p(\bm{\theta}_{n})Q(\bm{\theta}_{n}^{\star}|\bm{\theta}_{n})} \bigg\},
\label{acceptprob}
\end{equation}
where $Q(\bm{\theta}_{n}^{\star}|\bm{\theta}_{n})=p(\bm{\theta}_{n}^{\star}|\bm{\theta}_{n})$, the conditional proposal density and vice versa. Calculating the $Q$ ratio is necessary because the  Langevin-gradients  are not symmetric at different steps in the chain. 

\begin{figure*}[htbp!]
\centering
   \includegraphics[width=175mm] {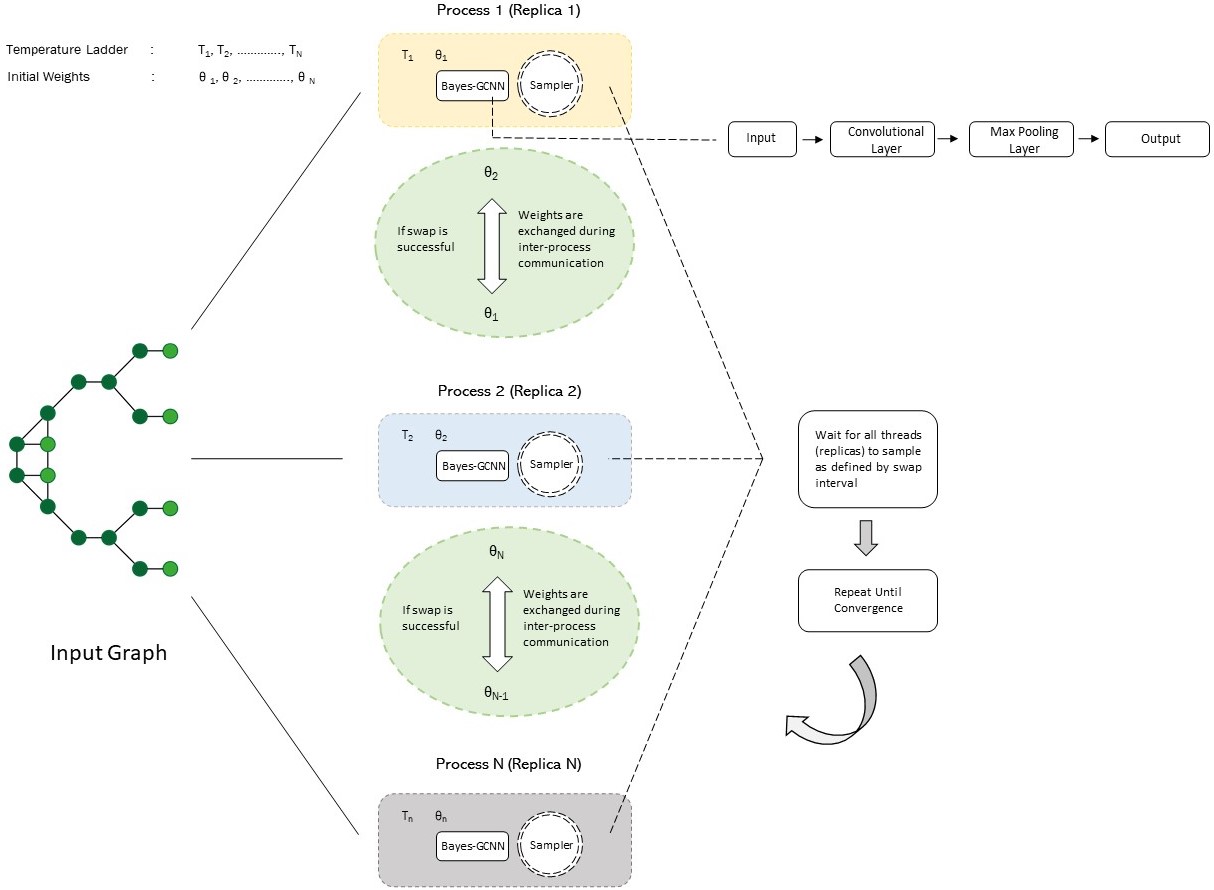}

\caption{ Bayesian GCNN framework implemented using  tempered MCMC and  parallel computing.}

\label{fig:framework}
\end{figure*}

Initially motivated by thermodynamics, tempered  MCMC (also known as replica exchange and parallel tempering MCMC) \cite{swendsen1986replica,hukushima1996exchange,hansmann1997parallel,swendsen1986replica}, samples an ensemble of $M$ MCMC replicas $\Omega = [R_1, R_2, .. R_M]$. Each replica features   temperature $t$ from the temperature ladder $T= [1, ... T_{max}]$ which is typically geometrically spaced, with $T_{max}$ defined by the user to control the extent of exploration. The likelihood of each replica in the ensemble is attenuated to form an attenuated posterior distribution $p_t(\bm{\theta}_{n}^t|\textbf{x})\propto p(\textbf{x}|\bm{\theta}_{n}^t)^{1/t}p(\bm{\theta}_{n}^t)$. This  ``flattens'' the distribution according to the temperature level of the replica, which increases the Metropolis--Hastings acceptance rates for replicas with higher temperature  levels, which also helps in escaping from local minima. The neighbouring replicas are periodically exchanged via a Metropolis--Hastings step, balancing exploration and exploitation  \cite{sen1996bayesian,maraschini2010monte}.  

Effectively, this augments the sample space of $\bm{\theta}_n$ into $\bm{\theta}_n^t=(\theta,\tau^2,t)$, sampling jointly across $\bm{\theta}$ and $t$. Those realisations with $t=1$ then have the target distribution. Therefore, periodically, a proposal is made to exchange the states of the neighbouring replicas, so that $\bm{\theta}_{n}^{t}=\bm{\theta}_{n}^{t+1}$ and $\bm{\theta}_{n}^{t+1}=\bm{\theta}_{n}^{t}$. The proposal distribution is symmetric and hence the priors cancel.  Therefore,  the acceptance probability is given by
\begin{equation}
\beta = \min\bigg\{1, \frac{P(\textbf{x}|\bm{\theta}_{n}^{t})^{1/(t+1)}P(\textbf{x}|\bm{\theta}_{n}^{(t+1)})^{1/t}}{P(\textbf{x}|\bm{\theta}_{n}^{(t+1)})^{1/(t+1)}P(\textbf{x}|\bm{\theta}_{n}^{t})^{1/t}} \bigg\}.\label{eq:PT}
\end{equation}.

\subsection{Bayes-GCNN framework}

\begin{figure}[htpb!]
 \centering
  \includegraphics[width=8cm]{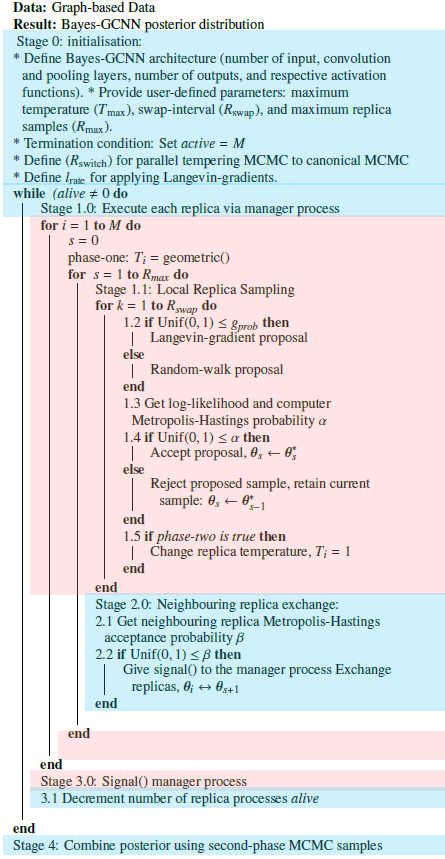}
 \caption{Tempered MCMC algorithm for Bayesian GCNN using parallel computing. Note that the Langevin-gradient proposal can be either  LG or adapt-LG. The manager process is highlighted in blue and replica processes running in parallel are highlighted in pink.}
\label{fig:Algorithm}
 \end{figure}

Our   Bayes-GCNN employs  parallel processing  for execution of tempered MCMC replicas that exchange states at regular interval via inter-process communication. The Bayes-GCNN framework that features parallel  MCMC replica samplers, inter-process communication,  and graph-based data is shown in Figure~\ref{fig:framework} and presented as an algorithm in Figure~\ref{fig:Algorithm}. We begin by defining the graph-CNN architecture that includes the size of convolution and max-pooling layers and number of output neurons given by the classification problem with given graph-based data representation for input as shown in Figure~\ref{fig:gcnn}.

 The execution begins with tempered  replica  sampling (Stage 1.1)  with  the manager process overseeing parallel replica processes. Each replica sample creates a proposal depending on the $l_{rate}$  (Stage 1.2) using either random-walk or  Langevin-gradients \eqref{update}. In  Stage 1.3,  the log-likelihood  is computed along with the Metropolis--Hastings probability  (Stage 1.4)  to accept/reject a proposed sample. Then, the algorithm checks (Stage 1.5) whether to carry on with tempered MCMC or  change the replica temperature values to 1 for canonical MCMC.  In this way, tempered MCMC is used in first-phase, then it switches to canonical MCMC in second-phase defined by $R_{\text{switch}}$ to further balance global exploration with local and to ensure that the true posterior distribution is sampled during the second-phase  as done in our previous works \cite{chandra2019langevin_,chandra2019PT-Bayeslands}. Hence, the tempered MCMC is used as a burn-in sampling procedure which is not part of the posterior distribution. The tempered MCMC is used for exploration but the samples do not become part of the true posterior, since it features pseudo-posterior distribution (due to the temperature level affecting the replica log-likelihood).

 Next the replica exchange is done depending on the replica swap-interval ($R_{\text{swap}}$) and   Metropolis-Hastings probability (Stage 2.2) which considers the log-likelihood of neighboring replica processes. We note that the manager process is used to determine if the neighboring replicas can be swapped. In case if they are swapped, the chain position is exchanged via inter-process communication. Finally, the algorithm decrements the number of active replicas  if the maximum number of replica samples ($R_{\text{max}}$) are reached which  enables the algorithm to end replica sampling. In post-replica sample stage (Stage 4),  burn-in sampling period is removed (which includes tempered MCMC)  and then combined with  respective replica posterior distribution of Bayes-GCNN for further analysis.

In addition to the configuration of the GCNN, the user sets the number of replicas ($M$), maximum temperature of the  temperature ladder  ($T_{\text{max}}$), neighbouring replica swap-interval ($R_{\text{swap}}$),  and maximum number of replica samples  ($R_{\text{max}}$). The user must also set the Langevin-gradient rate  ($l_{\text{rate}}$), which determines how often it is used  for creating the proposal, as opposed to a random-walk proposal (effectively setting $\nu_1=0$).

\section{Experiments and Results}
\label{sec:results}

We evaluate  the distinct features of Bayes-GCNNs in terms of computational efficiency of tempered  MCMC sampling, effect of  different  proposal distributions, and  prediction accuracy for established benchmark datasets. 

\subsection{Dataset description}

We use Cora  \cite{sen2008collective}, CiteSeer \cite{citeseer2003} and PubMed \cite{pubmed2012} citation network datasets, which are commonly used to evaluate   graph neural networks. Each dataset has one connected graph consisting of nodes  representing a scientific publication. The edges of the graph serve as citation links between the scientific publications (nodes). Each publication in each dataset is described by a word vector indicating the absence/presence of the corresponding word from the dictionary. The details of the dataset in terms of number of nodes, edges, classes and training and test samples are provided in Table~\ref{tab:data}.  We note that  number of training instances is  relatively low; however, this  specific data split is used in the literature \cite{Yang2016} and hence we used it for   comparing results. The data split  has 20 labels per class for training, and 1000 nodes for testing. %Ayush (check if this is correct)

The Cora dataset consists of citation information for 2708 machine Learning papers with  1433 unique words in its dictionary. The nodes are classified into 7 labels: ``case-based'', ``genetic algorithms'', ``neural networks'', ``probabilistic methods'', ``reinforcement learning'', ``rule learning'', and ``theory''. The CiteSeer dataset   extracted from the CiteSeer digital library consists of 3327 scientific papers with 3703 unique words in it's dictionary. The nodes are classified into 6 labels: ``agents'', ``artificial intelligence'', ``DB'', ``IR'', ``machine learning'', ``human computer interface''.  The PubMed dataset features papers about  diabetes which  contains 19717 scientific papers and 44,338 citation links with 500 unique words in the dictionary. The nodes are classified into 3 classes which include ``diabetes mellitus, experimental'', ``diabetes mellitus Type 1'' and ``diabetes mellitus Type 2''. 

\begin{table*}[htbp!]
\small
\center
\caption{Overview of datasets using  graph representation showing number (num.) of classes with training, and test  instances.}
 \begin{tabular}{l r r r r r} 
 \hline
 Dataset & Nodes & Edges &  Num. classes &   Num. training &   Num. testing \\ [0.5ex] 
 \hline   
 Cora & 2708 & 5429 & 7 & 140 & 1000 \\
 CiteSeer & 3327 & 4732 & 6 & 120 & 1000 \\
 PubMed & 19717 & 44338 & 3 & 60 & 1000 \\
 \hline
\end{tabular}
\label{tab:data}
\end{table*} 

\begin{table*}[htbp!]
\centering
\small
\caption{Bayes-GCNN topology showing the total number of parameters (weights and biases).}
\begin{tabular}{l r r r r r }
\hline
Dataset &  Input Neurons  &   Output Neurons   &  Hidden Layers & Total parameters\\
\hline
Cora & 1433 & 7 & 16 & 23063\\
CiteSeer & 3703 & 6 & 16 & 59366\\
PubMed & 500 & 3 & 16 & 8067\\
\hline
\end{tabular}
\label{tab:topology}
\end{table*}

\begin{table*}[htbp!]
\centering
\small 
\caption{Effect of  adaptive Langevin-gradient (adapt-LG) rate on the classification accuracy (acc.) for the Cora dataset. }
\begin{tabular}{r c c c  r r r }
\hline
adapt-LG Rate &Train Acc.(Mean, Max, Std)&Test Acc. (Mean, Max, Std) &Swap Per. &Accept Per. & Time (min.)\\
\hline
 0&  14.25 22.14 1.51
& 16.85	32.20 7.39
& 50.89
& 40.00
& 72.30
  \\
 0.25&  98.32 100.00 4.50
& 74.52	79.50 3.81
& 47.74
& 40.75
& 79.13
  \\
   0.5&  98.73 100.00 3.45
& 74.95	79.30 3.18
& 46.58
& 43.04 
& 83.43
  \\
 0.75&  98.94 100.00 2.28
& 75.35	79.30 2.17
& 47.04
& 45.25 
& 87.72
  \\
 1.0&  99.05 100.00 2.06
& 75.52	79.20 1.99
& 48.43
& 49.75
& 85.01
  \\
  
\hline
 
\end{tabular}
\label{tab:lgrate}

\end{table*}

\begin{table*}[htbp!]
\centering
\small 
\caption{Effect of number of replicas in tempered MCMC  for Cora dataset. }
\begin{tabular}{r c c  r r r}
\hline
\# Chains &Train Acc. (Mean, Max, Std) &Test Acc. (Mean, Max, Std) & Accept Per. &Swap Per. &Time (min.)\\
\hline
 2&  99.22 100.00 0.71
& 75.80 79.60 1.05
& 44.00 
& 43.62
& 138.64
  \\
 4&  99.06 100.00 2.01
& 75.45 79.20 1.84
& 43.75
& 46.34
& 84.24
  \\
 6&  98.97 100.00 2.33
& 75.38 79.30 2.19
& 44.00
& 45.89
& 82.82
  \\
 8&  98.89 100.00 2.50
& 75.11 78.80 2.50
& 44.38
& 47.04
& 84.76
  \\
  10&  98.70 100.00 3.39
& 74.92 79.30 3.13
& 43.80
& 47.52
& 87.23
  \\
  
  \hline

\end{tabular}
\label{tab:replicas}

\end{table*}

\subsection{Experiment Design}

We first run experiments to assess the effect of the tempered MCMC tuning parameters (hyper-parameters) and report the  computational time and classification accuracy. In all experiments, we use the following   parameters determined from  our trial experiments. In random-walk proposals, we create Gaussian noise with standard deviation $\nu_2=0.005$. We use maximum temperature ($T_{\text{max}}=2$) in tempered MCMC. We  change tempered MCMC to canonical MCMC  with $R_{\text{switch}}=60$ percent of total samples.  We use a swap interval $R_{\text{swap}}=2$  samples and  a maximum of  48,000 samples which are distributed across all the   replicas.

We implement Bayes-GCNN  using pyTorch\footnote{\url{https://pytorch.org/}} and pyTorch-geometric libraries\footnote{\url{https://pytorch-geometric.readthedocs.io/en/latest/}} and Python multi-processing library for parallel MCMC replica processes. Table~\ref{tab:topology} provides the topology of  Bayes-GCNN in terms of number of input, hidden, and output neurons for the respective datasets.

In the case of adaptive Langevin-gradients which is based on the Adam optimiser, the first and second order moments of the past gradient is used as a means of adapting the gradients   \cite{kingma2014adam}. In this case, we use a  user defined   learning rate based on the literature and trial experiments ($\nu_1=0.01$).  In the case of canonical Langevin-gradients,  we use user defined learning rate based on previous works  \cite{chandra2019langevin_,chandra2020bayesian,chandra2020surrogate}    ($\nu_1=0.1$). 
 
\subsection{Results}

We begin by evaluating  the effect of adaptive Langevin proposal (adapt-LG  rate)  on the proposal distribution  with 8 replicas in tempered MCMC using the Cora dataset.  Table~\ref{tab:lgrate} presents the   classification accuracy on training and testing datasets (showing mean, best, and standard deviation), neighbouring replica swap rate (percentage), and percentage of accepted samples during sampling.  We observe  a positive association between the adapt-LG rate and percentage of acceptance samples which implies   that adapt-LG provides better proposals when compared to  random-walk proposal distribution. There is also an association between adapt-LG rate and computational time since computing gradients is expensive \cite{chandra2019langevin_}. The accuracy for train and test dataset is much lower for adapt-LG rate of 0, which indicates that random-walk proposal distribution on its own cannot be used to train Bayes-GCNNs. However, higher adapt-LG rates do not appear to yield performance gains.

Table~\ref{tab:replicas} shows the  effect of number of replicas in tempered MCMC  for the Cora dataset according to the same metrics as used previously.  We observe  a reduction in computational time as the number of replicas increases from 2 to 4; however, the classification performance does not change much. This can be directly attributed to speed and efficiency gained by having more replica processes running in parallel (one replica process/thread per processing core). We also observe that there is a gradual  increase in the swap percentage (between neighboring replicas) for up to 8 replicas.

We then apply the Bayes-GCNN to the other datasets (CiteSeer and PubMed), using 8 replicas with adapt-LG rate of 0.75 and present the training and test classification accuracy  in Table~\ref{tab:lit}. The literature does not report the same result summaries (best, mean, standard deviation) that we do; hence, a direct comparison is not  possible. We compare  the results assuming those from literature as the best performance unless otherwise indicated. In the case of GCNN (SGD) \cite{Yang2016}, we report the mean classification accuracy. We also show performance by our own implementation of  canonical  GCNN* using Adam optimiser  with 30 independent experiential runs with different initial weights and biases. Note in case of Bayes-GCNN, the mean, best and standard deviation are taken from posterior distribution of one experimental run.

  Bayes-GCNN (adapt-LG) offers almost comparable  classification accuracy   to those in the literature (Table~\ref{tab:lit}). Although Bayes-GCNN (adapt-LG) offers slightly lower  classification accuracy (taking account the mean performance), it  provides comprehensive uncertainty quantification in predictions. We also note that using LG rather  than  adapt-LG proposal distribution significantly deteriorates the performance of Bayes-GCNN.

\begin{figure*}[htbp!]
    \centering
        \includegraphics[scale = 0.55]{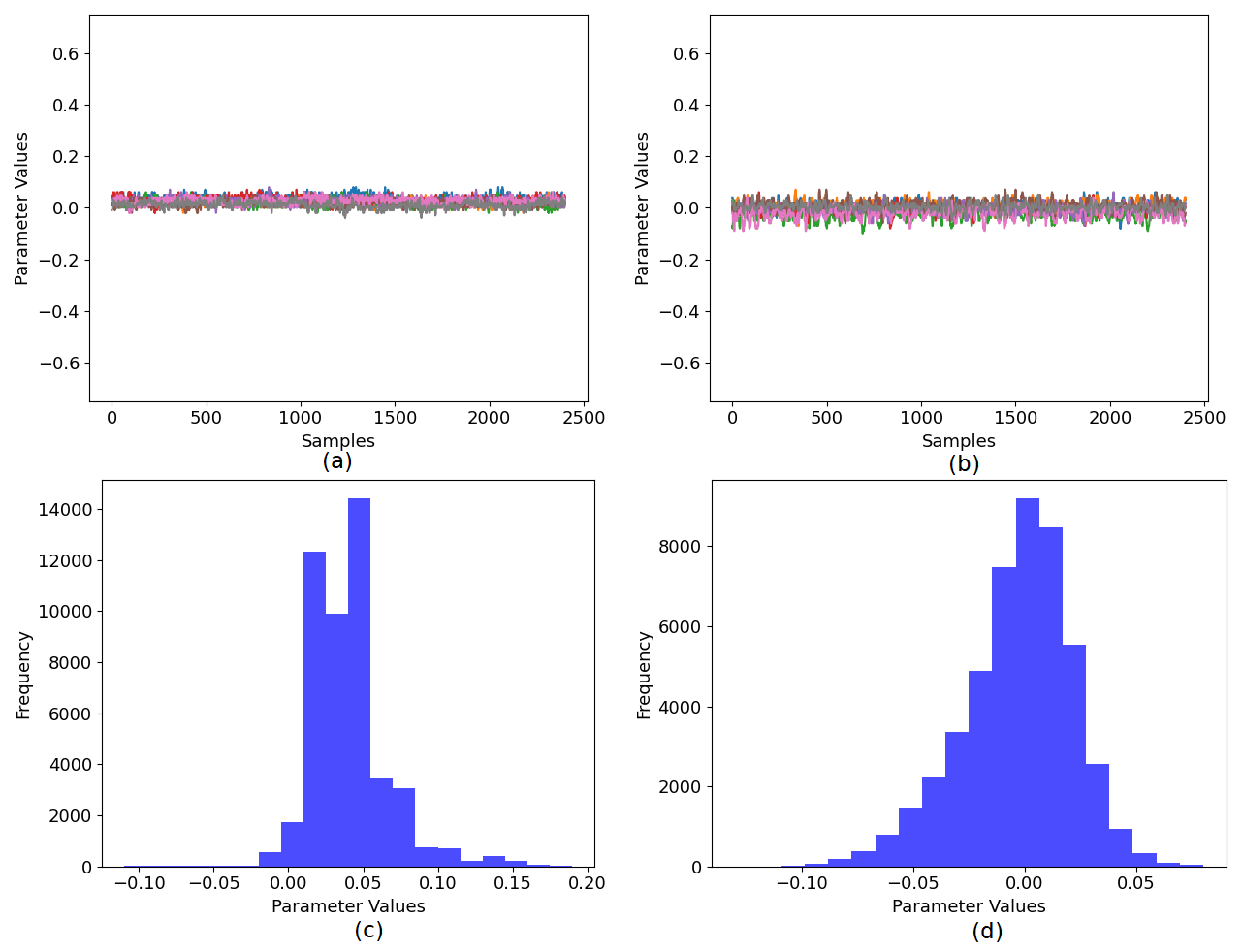}
    \caption{Posterior and trace plot for selected weights for Cora}
      \label{fig:posteriorCora}
\end{figure*}

\begin{figure*}[htbp!]
    \centering
        \includegraphics [scale = 0.55]{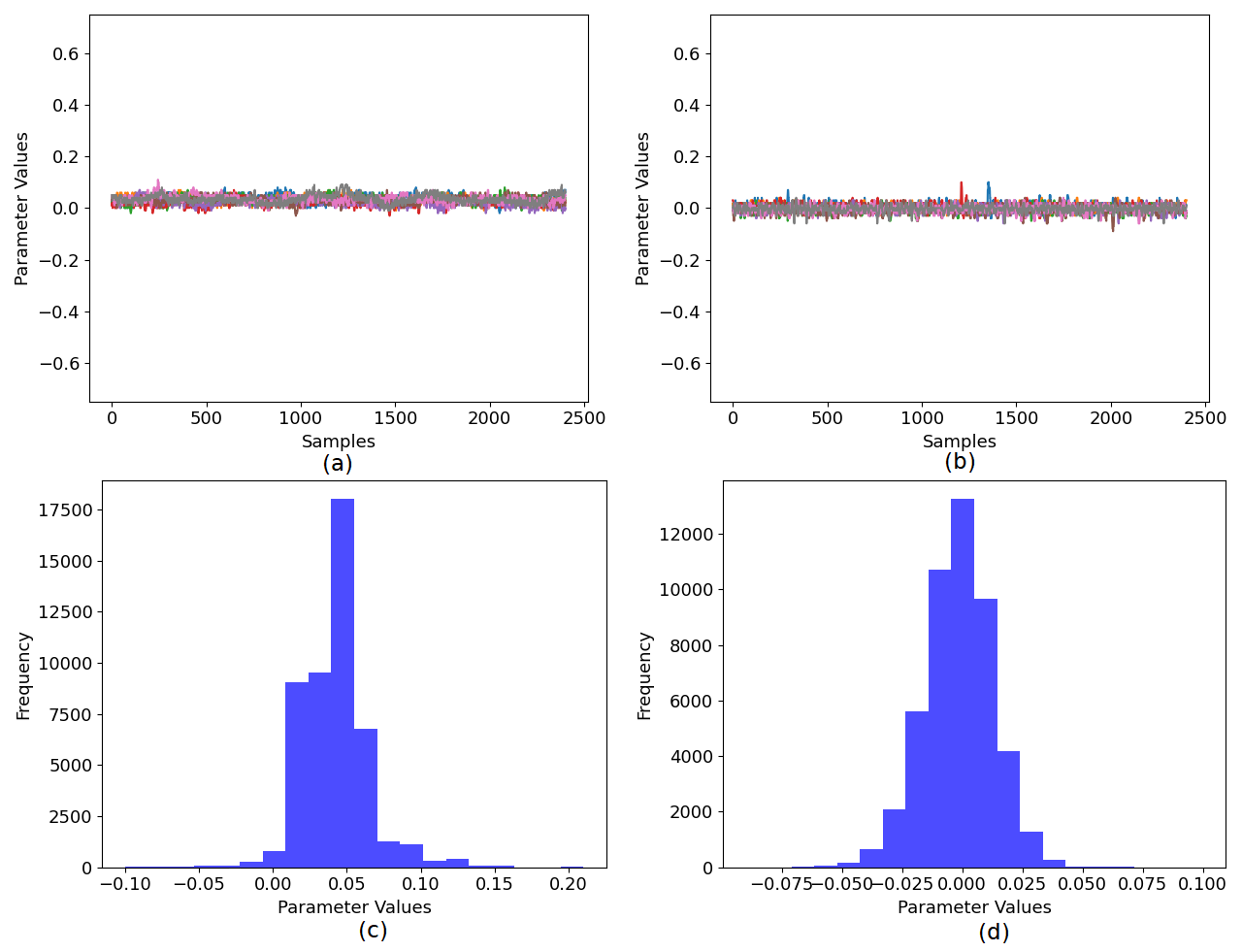}
    \caption{Posterior and trace plot for selected weights for CiteSeer}
      \label{fig:posteriorCiteSeer}
\end{figure*}

\begin{figure*}[htbp!]
    \centering
        \includegraphics[scale = 0.55]{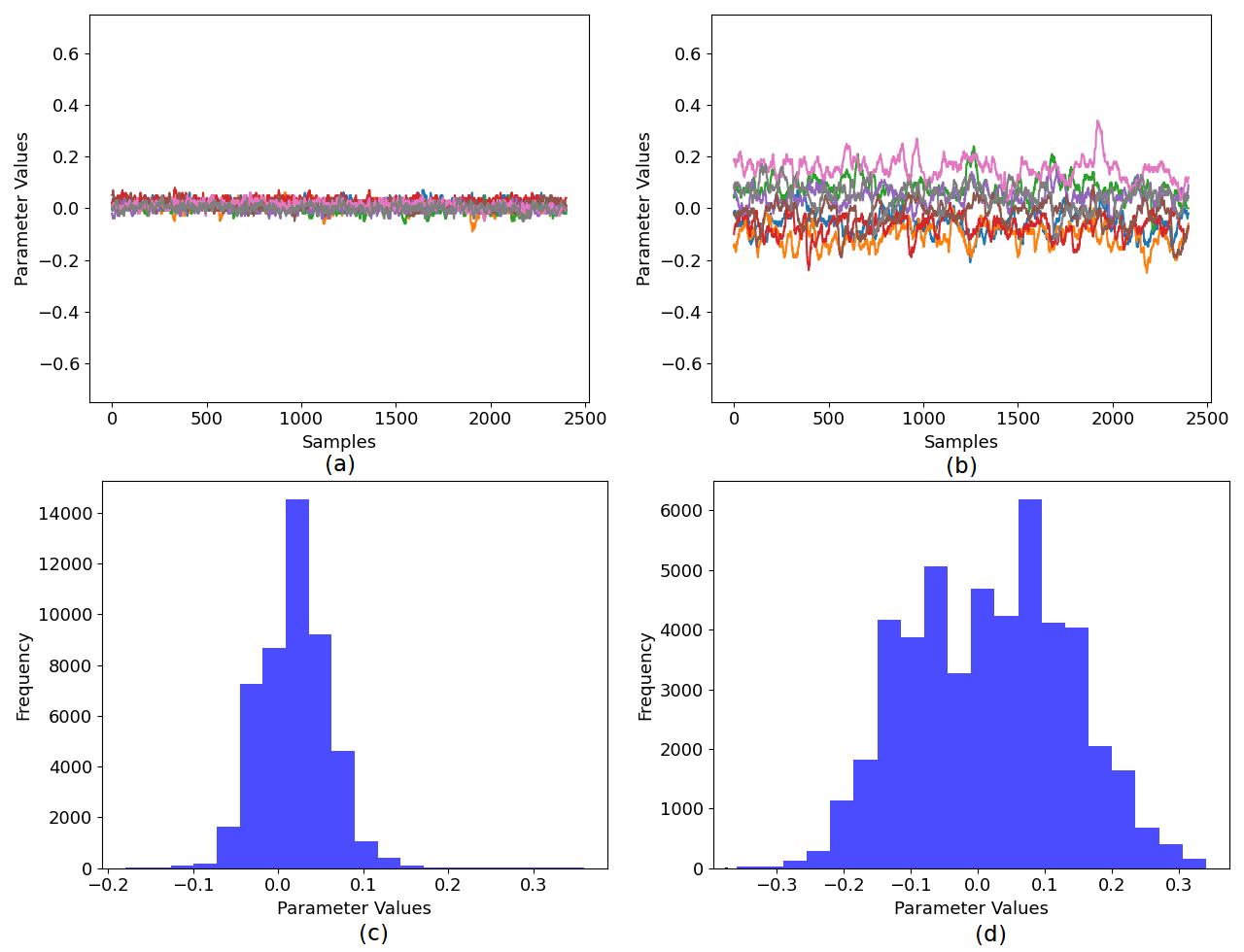}
    \caption{Posterior and trace plot for selected weights for PubMed}
      \label{fig:posteriorPubMed}
\end{figure*}

Figures~\ref{fig:posteriorCora}--\ref{fig:posteriorPubMed} show the Bayes-GCNN (adapt-LG) trace plot and posterior distribution for selected parameters (weights) from the respective  datasets. The trace plots show 8 replica samples with different colours post the burn-in period (hence all replica temperature values are 1).   
Figure~\ref{fig:posteriorCora} presents the  results from the Cora dataset showing the trace-plots (Panels a and c) and the posterior distribution (Panels b and d) for two selected weights  with evidence of a  unimodal posterior distribution. This can be seen in the dense histograms having a single peak for both the weights (Panels b and d). In the case for the CiteSeer  dataset shown in Figure~\ref{fig:posteriorCiteSeer}, we get a similar observation where the selected weights are   similar in trace plots (Panels a and c), and both the posterior distributions (Panels b and d) show a single peak and indicate a single node in a unimodal posterior. In Figure~\ref{fig:posteriorPubMed} (posterior for PubMed dataset), we see a striking  contrast between the trace plots (Panels a and c) of the selected weights: the  posterior distribution of the first weight (Panel b) indicates a  unimodal posterior, and the  posterior distribution of the second weight (Panel d) indicates a bimodal posterior. An explanation of this is that Cora and CiteSeer datasets employ a much larger Bayes-GCNN given by the number of parameters when compared PubMed. (See Table~\ref{tab:topology}.) All the trace plots show high correlation between the chains for all the respective datasets. 
 
The reason behind the difference in the trace-plot and respective posterior distributions between the different datasets cannot be explained without further analysis. We note that we  only selected two weights from thousands of parameters and this is merely for visualisation of the sampling process. The investigation as to why we get unimodal or multimodal posterior for different datasets and Bayes-GCNN architectures is beyond the scope of the paper.

\begin{figure*}[htbp!]
    \centering
        \includegraphics [scale = 0.70]{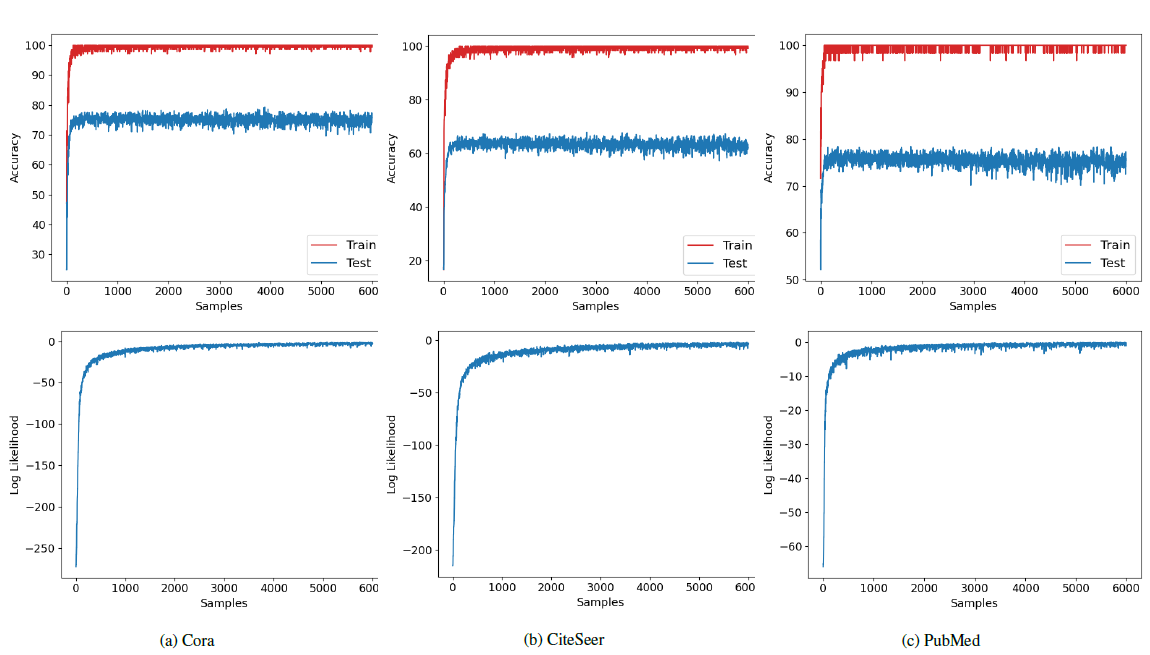}
    \caption{Accuracy And Log-Likelihood  for different problems}
      \label{fig:Accuracy And likelihood}
\end{figure*}

Figure~\ref{fig:Accuracy And likelihood} shows the log-likelihood plot along with the training and test classification accuracy for a single MCMC replica (with temperature level of 1) for   the respective problems. In  the respective  problems (Panel~a, b, and~c), we observe that the log-likelihood value increases in value over the time (samples), leading to higher  training and test  classification accuracy. We also notice that the case of PubMed dataset in Panel~c, is slightly different than Cora and CiteSeer datasets (Panel~a and~b). We find that PubMed dataset has a higher variance in test classification accuracy over time when compared to others. This could be purely due to the application problem and size of the datasets and the Bayes-GCNN architecture as shown in Table~\ref{tab:topology}.

Next, we present results to verify if the Bayes-GCNN (adapt-LG) has converged using the Gelman--Reubin diagnostic \cite{vats2020}. Table~\ref{tab:diagnostic} shows their values for selected weights for their respective problems;  $\hat{r}$ values close to 1 indicates convergence \cite{vats2020}.  We use the different chain replicas for selected weights with unique identity number (Weight-ID) to determine if there is convergence between the MCMC replicas. The diagnostic uses the posterior distribution from all the replica chains after a burn-in of 60\%. We observe that in  Table~\ref{tab:diagnostic}, all the selected weight-IDs for the respective problems have values close to 1 and hence have obtained convergence.

\begin{table*}

\small
\center
\caption{ Bayesian GCNN results comparison of established methods from the literature (unavailable values are left blank).}
 \begin{tabular}{l r r r } 
 \hline
 Method & Cora & CiteSeer & PubMed \\ 
 
  &best (mean, std) & best (mean, std) & best (mean, std)\\
 \hline 
   GCNN (SGD) \cite{Yang2016} 2016 & 75.70 (\hphantom{00.00}, \hphantom{00.0}) & 64.70 (\hphantom{00.00}, \hphantom{00.0})& 77.20 (\hphantom{00.00}, \hphantom{00.0})\\
 GCNN (Adam)  \cite{Kipf2016} 2017 & \hphantom{00.00} (81.50, \hphantom{00.0}) & \hphantom{00.00} (70.30, \hphantom{00.0}) & \hphantom{00.00} (79.00, \hphantom{00.0}) \\
 \hline
  %GCNN* (SGD) & 23.70 (16.78, 4.18)  & 25.60 (20.36, 2.74)  & 45.60 (38.13, 8.55)  \\
   Bayes-GCNN (LG) & 54.70 (40.98, 5.04) & 43.00 (30.82, 4.45) & 69.50 (60.53, 3.01)  \\
  GCNN* (Adam) & 81.70 (80.81, 0.65)  & 72.00 (70.51, 0.85) & 79.50 (79.00, 0.60) \\
 Bayes-GCNN (adapt-LG) & 79.00 (74.71, 2.42) & 68.90 (63.26, 2.74 ) & 78.70 (74.94, 1.63) \\
 \hline
\end{tabular}
\label{tab:lit}
\end{table*}

 \begin{table}[htbp!]
\centering
\small
 \caption{Convergence diagnostics for the selected weights denoted by identity number (Weight-ID) for respective problems (Cora, CiteSeer, and PubMed)}
\label{tab:diagnostic}
\begin{tabular}{ r r r r }
\hline
Weight-ID & Cora & CiteSeer & PubMed\\
\hline

0
& 1.25 & 1.27 & 1.25
\\
100
& 1.21 & 1.18 & 1.15
\\
1000
& 1.15 & 1.16 & 1.23
\\
5000
& 1.18 & 1.26 & 1.17
\\
8000
& 1.22 & 1.28 & 1.20
\\
\hline
\end{tabular}
\end{table}

\section{Discussion}
\label{sec:discussion}
 
 This paper serves as a proof-of-concept of implementing Bayesian inference via tempered MCMC for  GCNNs, achieving comparable performance   with traditional methods  (Table~\ref{tab:lit}). Bayes-GCNNs provide a principled approach to uncertainty quantification  for deep learning models. We observed that the adapt-LG proposal distribution performed significantly better than LG proposal distribution in Bayes-GCNN. This can be attributed to the fact that LG uses a single and constant learning rate for all the weights whereas adapt-LG  optimiser transforms the current gradient based on first and second moment of the past gradient. Hence, adapt-LG is better suited for convergence of large number of parameters when compared to  conventional LG proposal distribution. Since a single and constant learning rate is used for all the weights in LG, only a portion of the weights may reach their local minima; however, some of the  weights may not converge leading to poor performance or   inability to train as shown in Table~\ref{tab:lit}. 
 
The adaptive nature adapt-LG based on the Adam optimiser may also be a mild violation of the Markov assumption underpinning MCMC, since the exact step length depends on the previous gradient and the $Q$-ratio might be approximate; these requirements can be relaxed somewhat, but need to be checked carefully \cite{roberts_rosenthal_2007}, which can be the  subject for future work.

The Gelman--Rubin diagnostics  has been the most commonly used method to evaluate convergence of Markov chains, due to   ease of implementation and availability in software packages. However, it has also been reported to sometimes give a premature and unreliable convergence diagnosis, particularly in cases where the Markov chains are stuck in a local maxima \cite{Flegal_2008}. Moreover, the effectiveness of Gelman--Rubin diagnostics for large numbers of parameters is not well studied in the literature. Better  convergence diagnosis methods needs to be developed, since deep learning models features tens of thousands of parameters.  Moreover, auto-correlation and effective sample size has also been widely used for MCMC convergence \cite{roy2019convergence, auto2}, and their applicability for Bayesian deep learning models can also be evaluated.

The comparison of results with the literature motivates the implementation of Bayesian framework for other deep learning models, which includes LSTM and CNN models \cite{DuvenaudMAGHAA15}.  The uncertainty quantification in predictions can be useful for application areas, such as traffic forecasting \cite{BOGAERTS202062} and emotion recognition \cite{YIN2021106954}. We addressed graph-based CNNs in this paper; however, the approach can be used for other graph deep learning architectures such as graph-LSTM networks \cite{BOGAERTS202062,YIN2021106954} and conventional   graph neural networks \cite{DONON2020106547}.

The use of Bayesian inference via MCMC schemes in graph neural networks is largely unexplored. Future work could focus on implementing Bayesian inference via MCMC for other architectures  of graph neural networks, such as graph LSTM models\cite{BOGAERTS202062} and graph attention networks\cite{gan2018} using  different datasets (such as Quantum Machine 9)\cite{ramakrishnan2014quantum,doi:10.1021/ci300415d}, bioinformatics (enzymes) \cite{articleENZYME, articleBRENDA}, and social news network (Reddit) \cite{10.1145/2783258.2783417}. Further work could also focus on the efficacy of other types of MCMC schemes such as Hamiltonian MCMC methods\cite{betancourt2018conceptual}  to further improve the classification   performance.

\section{Conclusion}
\label{sec:conclusion}

 We presented Bayes-GCNN that featured tempered  MCMC sampling via parallel computing with adaptive Langevin-gradient proposal distribution. Our results indicated that while the mean accuracy of the Bayes-GCNN was around 4-5\% lower for the CiteSeer problem, the maximum accuracy in general is on par with the accuracy of canonical GCNN for all the benchmark problems. Hence, Bayes-GCNN provides an alternative form of training that features a principled approach to quantify uncertainty in model parameters. Bayes-GCNN eliminates the need to run repetitive experiments with a probabilistic representation of weights and biases. In addition, canonical optimisers do not offer uncertainty quantification on their own which is needed for certain problems; hence, Bayes-GCNN has good potential for real-world applications.
 
\section*{Code and Data}

Python-based implementation for  Bayes-GCNN  along with  data is available\footnote{\url{https://github.com/sydney-machine-learning/BayesianGraphNeuralNetworks}}.

\bibliographystyle{IEEEtran}
\bibliography{2018,Chandra-Rohitash,Bays,sample,cnnbayes,gnn2020,graphnn}

\EOD

\end{document}